\documentclass{article}

\usepackage[final]{corl_2018}

\usepackage{type1cm}
\usepackage[bottom]{footmisc}
\usepackage{graphicx,multicol,url}
\usepackage[para,online,flushleft]{threeparttable}
\usepackage{booktabs}
\usepackage{amssymb}
\usepackage{lmodern}
\usepackage{mathtools, nccmath}

\title{Online Continual Learning on Sequences}

\author{German I. Parisi$^{1,2}$ and Vincenzo Lomonaco$^{1,3}$\\
  $^1$ContinualAI Research\\
  $^2$University of Hamburg, Germany\\
  $^3$University of Bologna, Italy\\
  \url{{german.parisi, vincenzo.lomonaco}@continualai.org}\\
}

\begin{document}
\maketitle


\begin{abstract}
Online continual learning (OCL) refers to the ability of a system to learn over time from a continuous stream of data without having to revisit previously encountered training samples.
Learning continually in a single data pass is crucial for agents and robots operating in changing environments and required to acquire, fine-tune, and transfer increasingly complex representations from non-i.i.d. input distributions.
Machine learning models that address OCL must alleviate \textit{catastrophic forgetting} in which hidden representations are disrupted or completely overwritten when learning from streams of novel input.

In this chapter, we summarize and discuss recent deep learning models that address OCL on sequential input through the use (and combination) of synaptic regularization, structural plasticity, and experience replay.
Different implementations of replay have been proposed that alleviate catastrophic forgetting in connectionists architectures via the re-occurrence of (latent representations of) input sequences and that functionally resemble mechanisms of hippocampal replay in the mammalian brain.
Empirical evidence shows that architectures endowed with experience replay typically outperform architectures without in (online) incremental learning tasks.
\end{abstract}

\keywords{Continual learning, online learning, streaming learning, catastrophic forgetting, experience replay} 


\section{Introduction}
\label{sec:intro}

Real-world data is naturally non-stationary and temporally correlated.
Artificial learning systems, agents, and robots should thus be able to learn in a continual fashion from rich streams of non-i.d.d. input.
The ability of a system to continually acquire and fine-tune knowledge is known as continual learning (CL; see \cite{Chen18, Parisi18review} for recent reviews).
This paradigm is also referred to as lifelong learning (LL) in the literature.
Although CL and LL are arguably different (e.g. LL assumes a finite learning phase or a lifetime that ends at a given time, whereas CL is not necessarily subject to this constrain), these terms are mostly used interchangeably.
Machine learning models that learn in a continual fashion from non-stationary sequential input are of great interest to both the scientific community and the industry.
Firstly, CL should be a key property of large-scale systems to keep learning over time after their deployment with the goal to efficiently fine-tune and transfer knowledge and skills from continuous (and potentially infinite) streams of novel input.
Secondly, the development and evaluation of CL models will help provide valuable insights into how biological mechanisms of learning and memorization work in the brain~\cite{Aimone2009, Deng2010}.

Empirical evidence shows that connectionists architectures are prone to \textit{catastrophic forgetting}, i.e., when learning a new class or task, the overall performance on previously learned classes and tasks may abruptly decrease due to the novel input interfering with or completely overwriting existing representations~\cite{French1999, Mermillod13}.
Because catastrophic forgetting is a phenomenon that affects also deep learning models, the interest in CL models has grown almost exponentially in the machine learning community.
To prevent, or at least alleviate, catastrophic forgetting in neural networks, researchers have studied how to address the plasticity-stability dilemma~\cite{Grossberg1980}, i.e., how plastic should the models be to accommodate novel knowledge while preventing previously acquired knowledge to be forgotten and, thus, providing a certain degree of stability during the learning process.

The vast majority of CL models have taken inspiration from biological mechanisms of continual learning and can be divided into three main categories~\cite{Parisi18review}: synaptic regularization, structural plasticity, and memory replay.
Approaches using synaptic regularization impose additional constraints on the update of neural weights to protect consolidated knowledge~(e.g.~\cite{Kirkpatrick17, Zenke17}).
Synaptic regularization is inspired by theoretical neuroscience models
in which consolidated knowledge is protected from forgetting via synapses with a cascade of states yielding different levels of plasticity~\cite{Fusi2005}.
However, it remains unclear how to better implement it in artificial networks and, in particular, how to efficiently select the weights to be protected.
Approaches with structural plasticity apply architectural changes to a model, typically expanding it with additional neural resources to accommodate representations from novel input~(e.g.~\cite{Parisi2017a, Parisi18, Rusu16progressive})
For large-scale datasets, these models may have scalability issues and require additional modulatory mechanisms that control their growth over time.
Approaches with memory replay implement the re-occurrence of raw stored, training samples or latent representations of the input to prevent catastrophic forgetting~(e.g.~\cite{kemker18fearnet, Parisi18, pellegrini2019latent}).
Intuitively, storing previously processed input samples to be subsequently replayed to a model (a process known as rehearsal) can be a very expensive and prohibitive practice.
Instead, it has been shown to be more efficient to replay latent representations of the input, a process known as latent replay.
This latter type of replay is closer to biological mechanisms of hippocampal replay in the mammalian brain~\cite{Deng2010,Mcclelland95}.

\begin{figure}[t]
  \centering
  \includegraphics[width=0.5\textwidth]{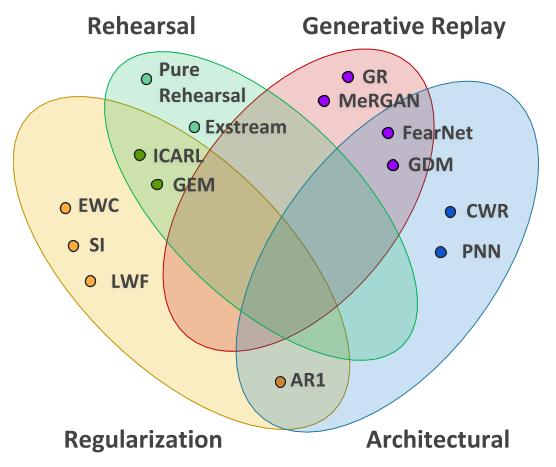}
  \caption{Venn diagram of some of the most popular CL strategies: CWR \cite{Lomonaco17}, PNN \cite{Rusu16progressive}, EWC \cite{Kirkpatrick17}, SI \cite{Zenke17}, LWF \cite{Li17}, ICARL \cite{Rebuffi16}, GEM \cite{Lopez-Paz17}, FearNet \cite{kemker18fearnet}, GDM \cite{Parisi18}, ExStream \cite{Hayes18MemoryEfficient}, Pure Rehearsal, GR \cite{Shin17}, MeRGAN \cite{wu2018memory} and AR1 \cite{Maltoni18}. Rehearsal and Generative Replay upper categories can be seen as a subset of replay strategies.}
  \label{fig:Venn}
\end{figure}

We show in Fig.~\ref{fig:Venn} the most popular and recent CL approaches divided into the above-described categories and their combinations.
In the diagram, we differentiate methods with \textit{rehearsal} (replay of explicitly stored training samples) from methods with \textit{generative replay} (replay of latent representations or the training samples).
Crucially, although an increasing number of methods have been proposed, there is no consensus on which training schemes and performance metrics are better to evaluate CL models.
Different sets of metrics have been proposed to evaluate CL performance on supervised and unsupervised learning tasks~(e.g.~\cite{Hayes18NewMetrics,Kemker17,Diaz18}).
In the absence of standardized metrics and evaluation schemes, it is unclear what it means to endow a method with CL capabilities.
In particular, a number of CL models still require large computational and memory resources that hinder their ability to learn in real time, or with a reasonable latency, from data streams.

Online CL (OCL) adds a number of desiderata to CL systems and their evaluation to ensure their ability to learn continually from potentially infinite sequential data in an online fashion, i.e., the ability to efficiently learn from one pass of the data.
In the next sections, we first define CL and OCL and then summarize and discuss recent deep learning models that address OCL on data streams through the use and combination of synaptic regularization, structural plasticity, and experience replay.

\section{Online Continual Learning}
\label{sec:ocl}

OCL builds on top of CL with a set of additional desiderata.
In particular, \textit{online} learning needs to take place as input data arrive and, since it should take place in real time, computational and memory resources must be considered.
Online learning shares a number of key properties with \textit{streaming} learning~\cite{Hayes18MemoryEfficient}, and the boundaries between these two paradigms get fuzzy in the machine learning literature.
In streaming learning, models are incrementally trained with one pattern at the time.
However, it is computationally more efficient to let OCL methods process the input also as mini-batches if the data is available~\cite{pellegrini2019latent}.
In this section, we first define CL and then discuss a set of desiderata for OCL on data sequences.

\subsection{\textbf{Formalizing CL Algorithms}}

We define CL, and thus OCL, within the framework presented in \cite{lesort:hal-02381343}: we assume CL aims to tackle a probably approximately correct (PAC) learnable problem in the approximation of a target hypothesis $h^*$ as well as learning from a sequence of non-i.i.d. training sets. This framework can also be seen as a generalization of the one proposed in \cite{Lopez-Paz17}, where learning happens continuously through a \textit{continuum} of data and a task supervised signal $t$ may be provided along with each training example.

In CL, data can be conveniently seen as drawn from a sequence of distributions $D_i$.
Let us define $\mathcal{D}$ is a potentially infinite sequence of unknown distributions $\mathcal{D} = \{D_1, \dots, D_N\}$ over $X \times Y$, where $X$ and $Y$ are input and output random variables respectively. At time $i$, a training set $Tr_i$ containing one or more observations is provided from $D_i$ to the algorithm.
A task is a learning experience characterized by a unique task label $t$ and its target function $g_{\hat{t}}^*(x) \equiv h^*(x,t=\hat{t})$, i.e., the objective of its learning.
It is important to note that tasks are an abstract representation of a learning experience carrying a task label.
This label helps split the full learning experience into smaller learning pieces. However, there is necessarily no bijective correspondence between data distributions and tasks.

Given $h^*$ as the general target function, i.e., our ideal prediction model, %
and a task label $t$, %
a CL algorithm $A^{CL}$ is an algorithm with the following signature: 
\begin{equation}
	\forall D_i \in \mathcal{D}, \hspace{20pt} A^{CL}_i:\ \ <h_{i-1}, Tr_i, M_{i-1}, t_i>  \rightarrow <h_i, M_i> 
\end{equation}

where:
\begin{itemize}
	\item $h_i$ is the current hypothesis at timestep $i$ (the parametric model learned continually);
	\item $M_i$ is an external memory where we can store previous training examples or partial computation not directly related to the parametrization of the model;
	\item $t_i$ is a task label that can be used to disentangle tasks and customize the hypothesis parameters. For simplicity, we can assume $N$ as the number of tasks, one for each $Tr_i$;
    \item $Tr_i$ is the training set of examples. %
    Each $Tr_i$ is composed of a number of examples $e_j^i$ with $j \in [1,\dots,m]$. Each example $e^{i}_j = <x^{i}_j, y^{i}_j>$, where $y^{i}$ is the feedback signal and can be the optimal hypothesis $h^*(x,  t)$ (i.e., exact label $y^{i}_j$ in supervised learning), or any real tensor (from which we can estimate $h^*(x, t)$, such as a reward $r^{i}_j$ in reinforcement learning). 
\end{itemize}

It is worth pointing out that each $D_i$ can be considered as a stationary distribution.
However, this framework allows to accommodate CL approaches where examples can also be assumed to be drawn in a non-i.i.d. fashion from each $D_i$ over $X \times Y$~\cite{Gepperth16,Hayes18NewMetrics}.

A CL scenario is a specific CL setting in which the sequence of $N$ task labels respects a certain task structure over time. Based on the proposed framework, we can define three different common scenarios:
\begin{itemize}
	\item \textsf{Single-Incremental-Task (SIT)}: $t_1 = t_2 = \dots = t_N $.
    \item \textsf{Multi-Task (MT)}: $ \forall i,j \in [1,.., n]^2, i \neq j, \implies t_i \neq t_j$.
    \item \textsf{Multi-Incremental-Task (MIT)}: $\exists\ i,j,k:\ t_i = t_j$ and $t_j \neq t_k $.
\end{itemize}

An example of a single-incremental-task (SIT) scenario is a classification task between cats and dogs, 
where the class distribution changes over time. First, there may only be input images of white dogs and white cats, whereas later only black dogs and black cats. Therefore, while learning to distinguish black cats from black dogs the algorithm should not forget to differentiate white cats from white dogs. The task stays the same but such a concept drift might lead to forgetting.
Instead, a multi-task (MT) scenario in a classification setting would first consist of learning cats versus dogs, and later cars versus bikes, without forgetting. The task label changes when the classes change and the algorithm can use this information to maximize its performance over time.
The multi-incremental-task (MIT) is the scenario where the same task can occur multiple times in the sequence of tasks, but such a task is not the only existing one.

\subsection{OCL on Sequences}
\label{sec:seq-learn}

OCL requires the quick integration of information during the learning process.
While typically supervised, unsupervised, and reinforcement learning paradigms divide the training phase from the test phase (with possible validation steps), in OCL the learning is seamless and such distinction can be applied to the data (i.e. to use a test-set that evaluates the model's performance) but cannot be used to constrain the underlying learning mechanisms.
In ecological learning environments, the data stream is assumed to be non-stationary and temporally correlated.
In this context, we can introduce four desiderata that better reflect the ability to continually learn from non-i.d.d. streams in an online fashion:
\\
\\
\textbf{1. Sequential Data}: We assume a potentially infinite data stream to be high-dimensional, non-stationary and temporally correlated. In OCL, incoming patterns must be processed one by one or as mini-batches for the quick integration of information while preventing catastrophic forgetting. Training strategies can take advantage of temporally correlated data streams to accelerate learning.
\\
\\
\textbf{2. Task-Agnosticism}: The model should work in the absence of supplementary supervised signals such as task boundaries or labels (i.e. $t_i=\emptyset$, $\forall D_i$).
In particular, the scenario in which an oracle provides the CL algorithm the task label both during training and testing to help reduce forgetting, disentangle representations and customize the agent behaviors constitutes a step-back with respect to the concept of CL algorithms which should learn continuously from a never-ending stream of data with none or very sparse supervised signals.
\\
\\
\textbf{3. Bounded Resources}: The number of training sets $Tr_i$ can potentially be unlimited, and thus, computation and memory should not be proportional to the number of hypothesis updates $h_i$ over time in any way.
While available computational and storage resources can significantly vary, a finite upper bound should rather exist and be considered, especially with $n \rightarrow \infty$.
Within the framework proposed in~\cite{lesort:hal-02381343}, the resource bound may be formulated as follows: for every step in time, the number of current examples contained in the memory is lower than the total number of previously seen examples:
    $\forall i \in [1,...,n], |M_i| \ll \biggl\lvert \bigcup\limits_{i=1}^{i-1} Tr_{i} \biggr\rvert$.
\\
\\
\textbf{4. Experience Replay}: The periodic replay of previously encountered data samples (e.g. stored in a replay buffer) can alleviate catastrophic forgetting.
However, storing data samples has the general drawback of large memory requirements and re-training complexity.
In the brain, hippocampal replay provides the means for the gradual integration of knowledge and is thought to occur through the re-activation of latent representations~\cite{Mcclelland95}
In latent replay, samples can be drawn from a probabilistic or generative model and replayed to the system for memory consolidation~\cite{Robins95}.

\subsection{Datasets and Benchmarks}
\label{sec:datasets}

Most of the proposed continual learning models have been designed for and evaluated with visual information.
Datasets such as \textsl{ImageNet} and \textsl{Pascal VOC} provide a very good playground for classification and detection approaches.
However, they were designed with static evaluation protocols in mind, i.e., the entire dataset is split into two parts: a training set is used for (one-shot) learning and a separate test set is used for performance evaluation.
Splitting the training set into a number of batches is essential to train and test continual learning approaches.

Most of the existing datasets are not well suited to this purpose because they lack a fundamental ingredient: the presence of multiple (unconstrained) views of the same objects taken in different video sessions (e.g. varying background, lighting, pose, occlusions).
The presence of temporally coherent sessions (i.e., videos where the objects move in front of the camera) is a key feature since temporal smoothness can be used to simplify object detection, improve classification accuracy, and address unsupervised scenarios~\cite{Maltoni2016-icpr}. This important propriety would indeed allow a natural interplay between sequence learning and CL approaches.

In the context of object recognition, we can consider three continual learning scenarios:
\begin{itemize}
\item \textbf{New Instances} (\textbf{NI}): new training patterns of the same classes become available in subsequent batches with new poses and conditions (illumination, background, occlusion, etc.). A CL model is expected to incrementally consolidate its knowledge about the known classes without compromising what it has learned before.
\item \textbf{New Classes} (\textbf{NC}): new training patterns belonging to different classes become available in subsequent batches. In this case, the model should be able to deal with the new classes without decreasing accuracy on the previous ones. 
\item \textbf{New Instances and Classes} (\textbf{NIC}): new training patterns belonging both to known and novel classes become available in subsequent training batches. A CL model is expected to consolidate its knowledge about the known classes and to learn the novel ones.
\end{itemize}

Most CL benchmarks are benchmarks adapted from others fields, for instance:
\begin{itemize}
  \item \textbf{Classification}: MNIST \cite{LeCun10}, Fashion-MNIST \cite{Xiao2017}, CIFAR10/100 \cite{Krizhevsky09}, Street View House Numbers (SVHN) \cite{Netzer11}, CUB200 \cite{Welinder10}, LSUN \cite{Yu15}, ImageNet \cite{krizhevsky12}, Omniglot \cite{lake2015human} {or Pascal VOC \cite{Everingham15} (object detection and segmentation)}.
  \item \textbf{Reinforcement Learning}: Arcade Learning Environment (ALE) \cite{Bellemare13} for Atari games, SURREAL \cite{Fan18} for robot manipulation and RoboTurk for robotic skill learning through imitation \cite{Mandlekar18}, \textit{CRLMaze} extension of VizDoom \cite{lomonaco2019continual} and DeepMind Lab \cite{Mankowitz18}. 
\end{itemize}

These datasets are then split, artificially modified via image rotations or the permutation of pixels, and concatenated together to create sequences of tasks. As an example, permuted MNIST \cite{Kirkpatrick17} and rotated MNIST \cite{Lopez-Paz17} are CL datasets artificially created from MNIST.

\begin{table}[t]
  \caption{Comparison of datasets (with temporal coherent sessions) for continual object recognition. * Temporal coherent training/test sessions for NORB and COIL-100 have been defined in \cite{Maltoni2016-icpr}.}
  \label{tab:datasets}
  \centering
  \begin{threeparttable}
  \begin{tabular}{p{3.2cm}p{0.6cm}p{0.7cm}p{0.8cm}p{1.3cm}p{1.2cm}p{1.8cm}p{1.3cm}}
    \toprule
    Dataset & Cat.  & Obj. & Sess. & Frames per sess. & Format & Acquisition setting & Outdoor sessions\\
    \midrule
    NORB \cite{LeCun2004} & 5 & 25 & 20 & 20\tnote{*} & grayscale & turntable & no\\
    COIL-100 \cite{Nene1996} & - & 100 & 20 & 54\tnote{*} & RGB & turntable & no\\
	iLab-20M \cite{Borji2016} & 15 & 704 & - & - & RGB & turntable & no\\
	RGB-D \cite{Schwarz2015} & 51 & 300 & - & - & RGB-D & turntable & no\\
	BigBIRD \cite{bigbird} & - & 100 & - & - & RGB-D & turntable & no\\
	ALOI \cite{aloi} & - & 1000 & - & - & RGB & turntable & no\\
	BigBrother \cite{Franco2009} & - & 7 & 54 & $\sim$20 & RGB & wall cam. & no\\
	iCubWorld28 \cite{Pasquale2015a} & 7  & 28 & 4 & $\sim$150 & RGB & hand hold & no\\
	iCubWorld-Transf \cite{icub-transf} & 15 & 150 & 6 & $\sim$150 & RGB & hand hold & no\\
	OpenLORIS \cite{icub-transf} & 19 & 69 & 7 & 500 & RGB-D & moving cam. & no\\
	CORe50 \cite{Lomonaco17} & 10 & 50 & 11 & $\sim$300 & RGB-D & hand hold & yes (3)\\
    \bottomrule
  \end{tabular}
  \begin{tablenotes}
    \vspace{5pt}
	\end{tablenotes}
  \end{threeparttable}
\end{table}

In Table \ref{tab:datasets}, we compare datasets that we believe are better suited for CL tasks (mostly in the context of continuous object recognition such as CORe50 \cite{Lomonaco17}, OpenLORIS \cite{openloris} or iCub-Transformation \cite{icub-transf}). 
Datasets where temporal coherent sequences are not available (or cannot be generated from static frames) were excluded.
In the first group of datasets~(\textsl{NORB}, \textsl{COIL-100}, \textsl{iLAB20M}, \textsl{Washington RGB-D}, \textsl{BigBIRD}, \textsl{ALOI}), objects are positioned on turntables and acquisition is systematically controlled in term of pose/lighting. Neither complex backgrounds nor occlusions are present in these datasets. For \textsl{NORB} and \textsl{COIL-100} we defined in \cite{Maltoni2016-icpr} a number of exploration sequences that turn the native static benchmarks into continual learning tasks; \cite{Maltoni2016-icpr} also reports supervised and semi-supervised accuracy for the NI scenario. Exploration sequences can be generated for the other datasets in this group as well by randomly walking through adjacent static frames in the multivariate parameter space; however, the obtained sequences would remain quite unnatural.   
\textsl{BigBrother} dataset (see \cite{Lomonaco2016} \cite{Franco2009}) is an interesting incremental learning setup in the face recognition domain but copyright restrictions do not allow the public distribution of the dataset. 
Finally, \textsl{iCubWorld} (\cite{Pasquale2015a}, \cite{icub-transf}) and OpenLORIS have been acquired in a robotic vision context and are similar to \textsl{CORe50}. In fact, objects are handhold at nearly constant distance from the camera and are randomly moved. Among all these datasets, \textsl{CORe50} and \textsl{OpenLORIS} comprise a higher number of longer sessions (including outdoor ones), more complex backgrounds and also provide depth information that can be used as extra-feature for classification and/or to simplify object detection.
We think that cross-evaluating online continual learning approaches on \textsl{CORe50}\footnote{\url{vlomonaco.github.io/core50}}, \textsl{iCubWorld-Transf} and \emph{OpenLORIS} could be very interesting.

\begin{figure}[t]
  \centering
  \includegraphics[width=0.8\textwidth]{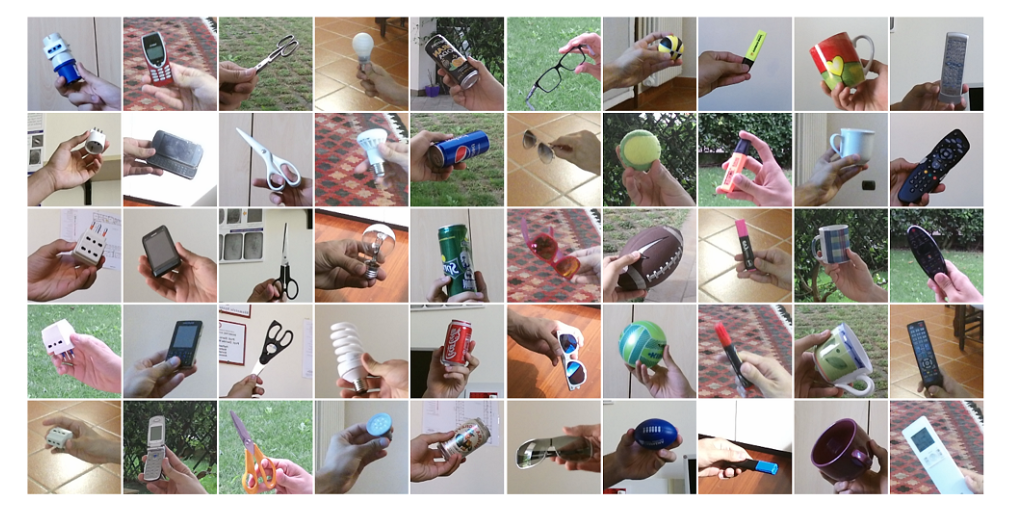}
  \caption{Example images of the 50 objects in \textsl{CORe50}. Each column denotes one of the 10 categories~\cite{Lomonaco17}.}
  \label{img:core50}
\end{figure}

\begin{figure}[t]
  \centering
  \includegraphics[width=\textwidth]{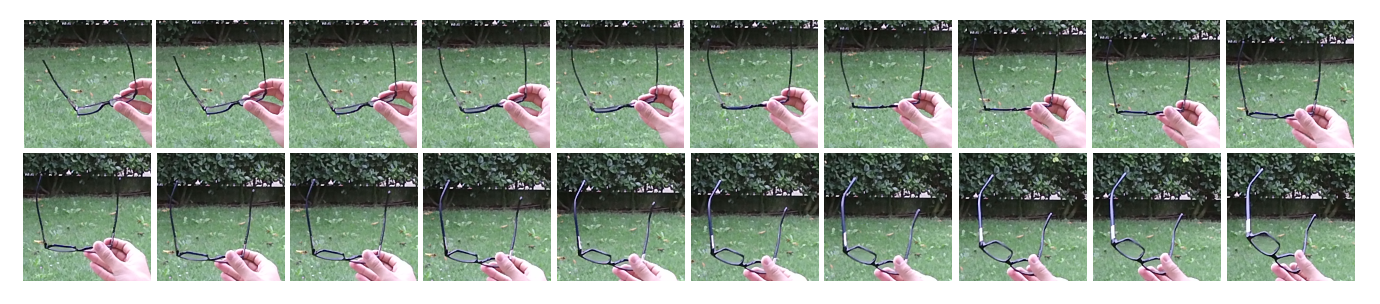}
  \caption{Example of 1 second recording (at 20 fps) of object \#26 in session \#4 (outdoor)~\cite{Lomonaco17}.}
  \label{img:seq}
\end{figure}

\section{Hybrid Approaches for Gradient-Based OCL}
\label{sec:regularization}

In this section, we will discuss a number of recently proposed algorithms for OCL based on gradient-descent. We will start with simple algorithms inspired by structural plasticity in the brain, moving towards more complex and hybrid approaches also using synaptic regularization as well as experience replay. We show that simply combining different and complementary approaches for OCL results in higher memory retention and, ultimately, in a better overall performance. 

\subsection{Copy Weight with Reinit (CWR)}
\label{sec:strategies}

CWR~\cite{pmlr-v78-lomonaco17a} was proposed as a baseline technique for CL from sequential batches. While this approach can work for NC (new classes) as well as for NIC (new instances and classes) update content type, here we focus on NC under SIT scenario.
The most obvious approach to implement an SIT strategy seems to be:
\begin{enumerate}
	\item Freeze shared weights $\bar{\Theta}$ after the first batch.
	\item For each batch $B_i$, extend the output layers with new neurons/weights for the new classes, randomly initialize the new weights but retain the optimal values for the old class weights. The old weights could then be frozen (denoted as FW in \cite{Lomonaco2017-arxiv}) or continued to be tuned (denoted as CW in \cite{Lomonaco2017-arxiv}).
\end{enumerate}

To learn class-specific weights without interference among batches, CWR maintains two sets of weights for the output classification layer: $cw$ are the consolidated weights used for inference and $tw$ the temporary weights used for training: $cw$ are initialized to 0 before the first batch, while $tw$ are randomly re-initialized (e.g., Gaussian initialization with std = 0.01, mean = 0) before each training batch.
In other words, $cw$ can be seen as a sort of hippocampus where consolidated concepts are maintained, while $tw$ as a short-term working memory in the cortex used to learn new concepts without interfering with stable ones.
At the end of each batch training, the weights in $tw$ corresponding to the classes in the current batch are scaled and copied in $cw$: this is trivial in NC case because of the class segregation in different batches but is also possible for more complex cases~\cite{pmlr-v78-lomonaco17a}.
To avoid forgetting in the lower levels, after the first batch $B_1$, all the lower level weights $\bar{\Theta}$ are frozen. Weight scaling with batch-specific weights $w_i$ is necessary in case of unbalanced batches with respect to the number of classes or number of examples per class.

In CWR experiments reported in \cite{pmlr-v78-lomonaco17a}, we used models without class-shared fully connected layers (e.g., we removed FC6 and FC7 in CaffeNet) to better disentangle class-specific weights.
Since $\bar{\Theta}$ weights are frozen after the first batch, fully connected layer weights tend to specialize in the first batch classes only.
CWR implementation is very simple and, the extra computation is negligible and for each batch $B_i$, its overhead consists of the storage of temporary weights $tw$, totaling $s \cdot pn$ values, where $s$ is the number of classes and $pn$ the number of penultimate layer neurons.

\paragraph{\textbf{Copy Weight with Reinit+ (CWR+)}}
\label{sub:mean_shift_and_zero_initialization_cwr_}

In \cite{maltoni2019}, the authors proposed two simple modifications of CWR: the resulting approach is denoted as CWR+. 
The first modification, mean-shift, is an automatic compensation of batch weights $w_i$.
Tuning such parameters is annoying and a wrong parametrization can lead the model to underperform.
Empirical evidence shows that if the weights $tw$ learned during batch $B_i$ are normalized by subtracting their global average, then rescaling by $w_i$ is no longer necessary (i.e., all $w_i = 1$).
Other reasonable forms or normalization, such as setting standard deviation to 1, led to worse results in our experiments. 
The second modification, denoted as zero init, consists in setting initial weights $tw$ to 0 instead of typical Gaussian or Xavier random initialization.
It is well known that neural network weights cannot be initialized to 0, because this would cause intermediate neuron activations to be 0, thus nullifying back-propagation effects.
While this is certainly true for intermediate level weights, it is not the case for the output level \cite{maltoni2019}.

What is important here is not using the value 0, but the same value for all the weights (0 is used for simplicity).
It has been shown that this has a significant impact on the training dynamic and the forgetting~\cite{maltoni2019}. If output level weights are initialized with Gaussian or Xavier random initialization they typically take small values around zero, but even with small values in the first training iterations the softmax normalization could produce strong predictions for wrong classes.
This would trigger unnecessary errors backpropagation changing weights more than necessary.
While this initial adjustment is irrelevant for normal batch training, we empirically found that is detrimental for continual learning.

\subsection{\textbf{Architect \& Regularize (AR1)}}

A drawback of simple structural plasticity approaches as CWR and CWR+ is that weights $\bar{\Theta}$ are tuned during the first batch and then frozen. AR1, firstly proposed in \cite{maltoni2019}, consists of the combination of an Architectural and Regularization approach. In particular, CWR+ is extended by allowing $\bar{\Theta}$ to be tuned across batches subject to a regularization constraint (as per LWF \cite{Li17}, EWC \cite{Kirkpatrick17} or SI \cite{Zenke17}). The authors performed several combination experiments on CORe50 to select a regularization approach; each approach required a new hyperparameter tuning w.r.t. the case when it was used in isolation. At the end, the choice for AR1 was in favor of SI \cite{Zenke17} because of the following reasons:

\begin{itemize}
	\item LWF performs nicely in isolation, but in our experiments it does not bring relevant contributions to CWR+. Being the LWF regularization driven by an output stability criterion, most of the regularization effects go to the output level that CWR+ manages apart. 
	\item Both EWC and SI provide positive contributions to CWR+ and their difference is minor. While SI can sometimes be unstable when operating in isolation, it has been shown that it is much more stable and easy to tune when combined with CWR+~\cite{maltoni2019}.
	\item SI overhead is small since the computation of trajectories can be easily implemented from data already computed by SGD.
\end{itemize}

Considering the low computational overhead and the fact that typically SGD is typically early stopped after 2 epochs, AR1 is suitable for online continual learning. 

\subsubsection*{AR1* and Latent Replay}

A simple approach such as CWR+, where the fully connected layer is implemented as a double memory, is quite effective to control forgetting in the SIT - NC scenario.
However, after the first training batch, CWR+ freezes all the layers except the last one, thus losing the benefit of an incremental adaptation of the underlying representation. AR1 \cite{maltoni2019} was then proposed to extend CWR+ by enabling end-to-end continual training throughout the entire network. To this purpose, the Synaptic Intelligence \cite{Zenke17} regularization approach (similar to EWC \cite{Kirkpatrick17}) is adopted to constrain the change of critical weights. In \cite{Lomonaco2019}, the authors:
\begin{enumerate}
    \itemsep0.01em 
	\item adapt CWR+ to the NIC scenario, thus making it able to reload past weights for already known classes and to adapt them with weighted contributions from different batches. As AR1 incorporates CWR+ in its main algorithm, this modification will result in two continual learning strategies, denoted as CWR* and AR1*.
	\item show that in a fine-grained scenario with small and non i.i.d. batches, Batch Normalization layers thwart the continual learning process and replacing them with Batch Renormalization \cite{loffe2017} can effectively tackle this problem.
	\item propose a selective weight freeze for the CNN models adopting Depth-Wise Separable Convolutions.
	\item reduce the computational and storage complexity of AR1 (and in general of EWC like approaches), by introducing an alternative way to implement weights update starting from the Fischer matrix.
\end{enumerate}

While 1. is specific to CWR+, 2-4. can be applied to other CL approaches as well.


AR1* proved to be effective even without any kind of experience replay.
However, even a small percentage of replay patterns can be extremely beneficial for taming forgetting~\cite{pellegrini2019latent}.
In deep neural networks, the layers close to the input (often referred to as representation layers) usually perform low-level feature extraction and, after the proper pre-training on a large dataset (e.g., ImageNet), their weights are quite stable and reusable across applications.
Higher layers, instead, tend to extract class-specific discriminant features and their fine-tuning is often important to maximize accuracy.

\begin{figure}[t]
\centering
\includegraphics[width=0.65\columnwidth]{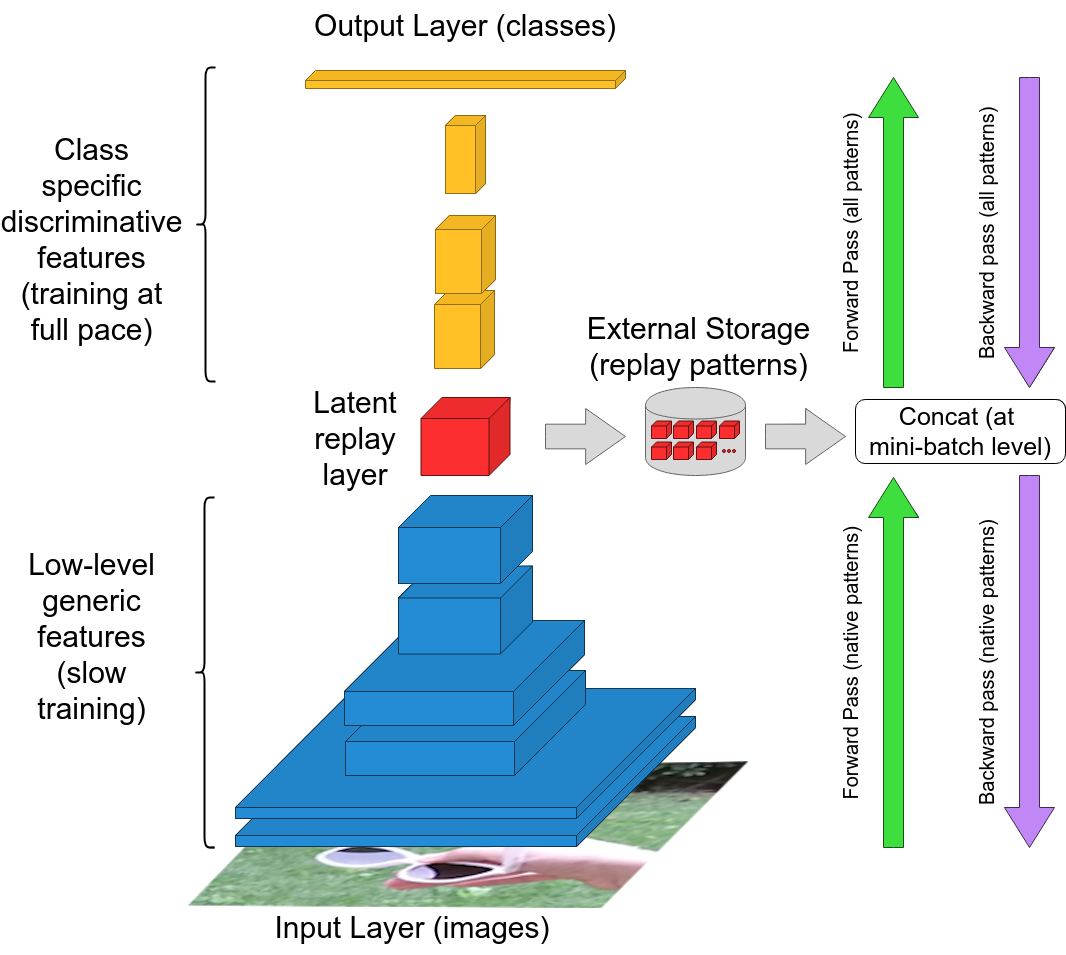}
\caption{Architectural diagram of Latent Replay~\cite{pellegrini2019latent}.}
\label{fig:latent_replay}
\end{figure}

For this reason, \emph{latent replay} was introduced~\cite{pellegrini2019latent}: instead of maintaining copies of input patterns in the external memory in the form of raw data, they stored the activations volumes at a given layer (denoted as \emph{Latent Replay layer}; see Fig.~\ref{fig:latent_replay}).
To keep stable representations and valid stored activations, it was proposed to slow down the learning at all the layers below the latent replay one and to leave the layers above free to learn at full pace.
In the limit case where low layers are completely frozen (i.e., slow down to 0), latent replay is functionally equivalent to rehearsal from the input (hereafter denoted as \emph{native rehearsal}), but achieves a computational and storage saving thanks to the smaller fraction of patterns that need to flow forward and backward across the entire network and the typical information compression that networks perform at higher layers.

\begin{figure}[th]
\centering
\includegraphics[width=0.65\columnwidth]{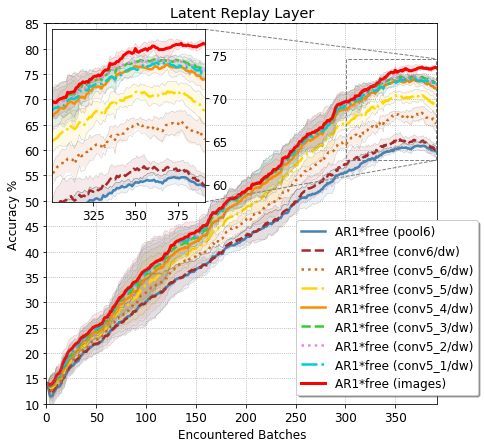}
\caption{AR1*free with latent replay ($RM_{size}=1500$) for different choices of the latent replay layer~\cite{pellegrini2019latent}. Setting the replay layer at the \texttt{pool6} layer makes AR1*free equivalent to CWR*. Setting the replay layer at the ``images'' layer corresponds to native rehearsal. The 
saturation effect which characterizes the last training batches is due to the data distribution in NICv2 -- 391 (see \cite{Lomonaco2019}): in particular, the lack of new instances for some classes (that already introduced all their data) slows down the accuracy trend and intensifies the effect of activations aging.}
\label{fig:latent_layer_diff}
\end{figure}

\begin{figure}[th!]
\centering
\includegraphics[width=0.6\columnwidth]{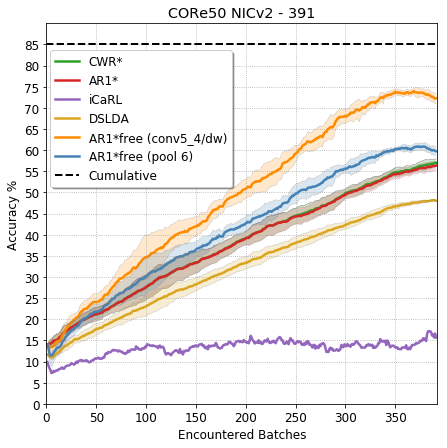}
\caption{Accuracy results on the CORe50 NICv2 -- 391 benchmark of CWR*, AR1*, DSLDA, iCaRL, AR1*free (\texttt{conv5\_4}), AR1*free (\texttt{pool6})~\cite{pellegrini2019latent}. Results are averaged across 10 runs in which the batches order is randomly shuffled. Colored areas indicate the standard deviation of each curve. As an exception, iCaRL was trained only on a single run given its extensive run time ($\sim$14 days).}
\label{fig:comparison}
\end{figure}

In the general case where the representation layers are not completely frozen the activations stored in the external memory suffer from an aging effect (i.e., over time they tend to increasingly deviate from the activations that the same pattern would produce if feed-forwarded from the input layer). However, if the training of these layers is sufficiently slow the aging effect is not disruptive since the external memory has enough time to be rejuvenated with fresh patterns. When latent replay is implemented with mini-batch SGD training: \emph{(i)} in the forward step, a concatenation is performed at the replay layer (on the mini-batch dimension) to join patterns coming from the input layer with activations coming from the external storage; \emph{(ii)} the backward step is stopped just before the replay layer for the replay patterns.

In Figure \ref{fig:latent_layer_diff}, we show the accuracy of AR1*free with latent replay ($RM_{size}=1500$) for different choices of the rehearsal layer (reported between parenthesis) is shown. As expected, when the replay layer is pushed down the corresponding accuracy increases, showing that a continual tuning of the representation layers is important. However, after \texttt{conv5\_4/dw} there is a sort of saturation and the model accuracy no longer improves. The residual gap ($\sim$4\%) with respect to native rehearsal is not due to the weights freezing of the lower part of the network but to the aging effect introduced above. This can be  proved by implementing an ``intermediate'' approach that always feeds the replay pattern from the input and stops the backward at \texttt{conv5\_4}: such an intermediate approach achieved an accuracy very close to the native rehearsal at the end of the training. The accuracy drop due to the aging effect can be further reduced with better tuning of BNR hyper-parameters and/or with the introduction of a scheduling policy, making the global moment mobile windows wider as the continual learning progresses (i.e., more plasticity in the early stages and more stability at later stages).
In Figure \ref{fig:comparison}, we show the iCaRL accuracy over time and compare it with AR1*free (\texttt{conv5\_4/dw}), AR1* (\texttt{pool6}) as well as the top three performing rehearsal-free strategies on CORe50 NICv2-391 (CWR*, AR1* and DSLDA) are reported. While iCaRL exhibits better performance than LWF and EWC (as reported in \cite{Lomonaco2019}), it is far from DSLDA, CWR* and AR1*.

Overall, AR1* combined with latent replay shows to be substantially more effective for OCL on sequential data streams with a good resource consumption trade-off~\cite{pellegrini2019latent}).

\section{Growing Networks with Experience Replay}
\label{sec:structural}

In this section, we introduce two approaches using growing neural networks for unsupervised learning and human action classification from videos.
The architectures discussed here can learn also in the absence of an explicit teaching signal such a class label.
The requirement for dense human annotations used by standard regression techniques is undesirable as such labels are typically not present in real-world learning scenarios.

The architectures comprise growing self-organizing networks for learning from sequential input.
Specifically for self-organizing networks, catastrophic interference is modulated by the conditions of map plasticity, the available resources to represent information, and the similarity between new and old knowledge~\cite{Parisi17, Richardson08}.
Growing networks can dynamically add new neurons and synapses to accommodate novel knowledge and protect consolidated embeddings from catastrophic interference.
Importantly, we will describe how self-organizing learning dynamics can be leveraged to implement efficient experience replay mechanisms without the need for additional memory resources such as ad-hoc replay buffers.

\subsection{Growing Self-Organizing Networks}

\begin{figure*}[t]
\centering
\includegraphics[width=0.85\textwidth]{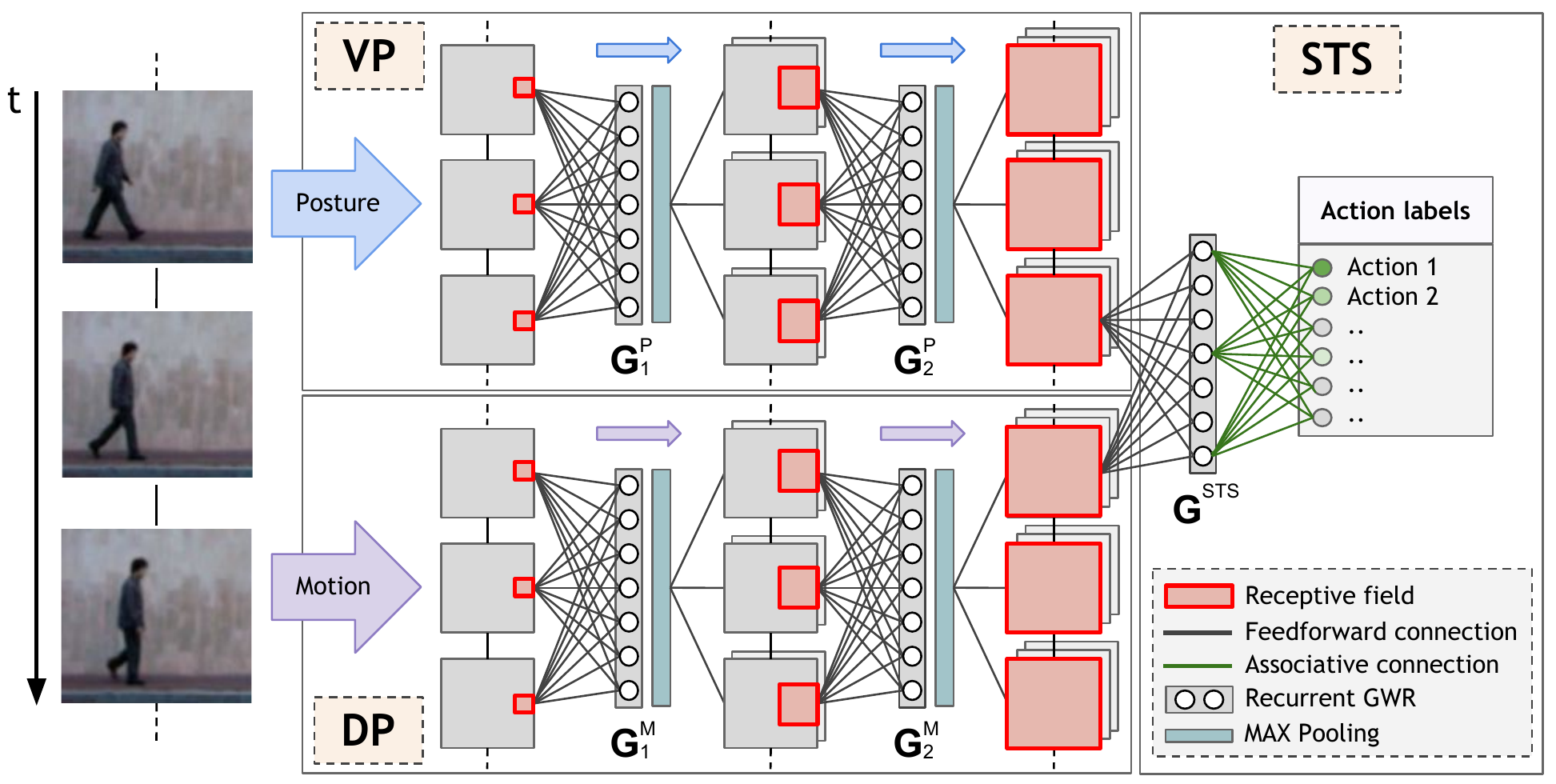}
\caption[Unsupervised deep learning architecture]{Diagram of our deep neural architecture with Gamma-GWR networks for continual action recognition~\cite{Parisi17}.}
\label{fig:dla}
\end{figure*}

In Parisi \textit{et al.}~\cite{Parisi17}, we proposed a self-organizing architecture consisting of a series of hierarchically arranged growing networks for the continual learning of human actions from videos.
Each layer in the hierarchy comprises a growing recurrent network Gamma-GWR and a pooling mechanism for learning action features with increasingly large spatiotemporal receptive fields~(Fig.~\ref{fig:dla}).
The proposed deep architecture is composed of two distinct processing streams for pose and motion features, and their subsequent integration in the STS layer.

The Gamma-GWR extends the Grow When Required (GWR) network~\cite{Marsland2002} with temporal context.
Each neuron consists of a weight vector $\textbf{w}_j$ and a number $K$ of context descriptors $\textbf{c}_{j,k}$~(with $\textbf{w}_j,\textbf{c}_{j,k}\in\mathbb{R}^n$).
Given $\textbf{x}(t)\in\mathbb{R}^n$ as input, the index of the best-matching unit (BMU), $b$, is computed as:
\begin{equation} \label{eq:GetB}
b = \arg\min_{j\in A}(d_j),
\end{equation}
\begin{equation} \label{eq:BMUr}
d_j = \alpha_0 \Vert \textbf{x}(t) - \textbf{w}_j  \Vert + \sum_{k=1}^{K}\ \alpha_k \Vert \textbf{C}_k(t)-\textbf{c}_{j,k}\Vert,
\end{equation}
\begin{equation}\label{eq:MergeStep}
\textbf{C}_{k}(t) = \beta \cdot \textbf{w}_b^{t-1}+(1-\beta) \cdot \textbf{c}_{b,k-1}^{t-1},
\end{equation}
where $\Vert \cdot \Vert$ denotes the Euclidean distance, $\alpha_i$ and $\beta$ are constant values that modulate the influence of the temporal context, $\textbf{w}_b^{t-1}$ is the weight vector of the BMU at $t-1$, and $\textbf{C}_{k}\in\mathbb{R}^n$ is the global context of the network with $\textbf{C}_{k}(t_0)=0$.
For an input $\textbf{x}(t)$, the activity of the network, $a(t)$, is defined in relation to the distance between the input and its BMU (Eq.~\ref{eq:GetB}):
\begin{equation} \label{eq:Activity}
a(t)=\exp(-d_b),
\end{equation}
thus yielding the highest activation value of $1$ when the network can perfectly match the input sequence~($d_b=0$).
The training of the neurons is carried out by adapting the BMU $b$ and its neighboring neurons $n$:
\begin{equation}\label{eq:UpdateRateW}
\Delta \textbf{w}_i = \epsilon_i \cdot h_i \cdot (\textbf{x}(t) - \textbf{w}_i),
\end{equation}
\begin{equation}\label{eq:UpdateRateC}
\Delta \textbf{c}_{i, k} = \epsilon_i \cdot h_i \cdot (\textbf{C}_k(t) - \textbf{c}_{i, k}),
\end{equation}
where $i\in\{b,n\}$, $\epsilon_i$ is a constant learning rate ($\epsilon_n<\epsilon_b$), and $h_i$ is a habituation counter that decreases over time as the neurons are fired (i.e., selected as the BMU or its neighbor).

Empirical studies have shown that Gamma-GWR networks with additive neurogenesis show a better performance than a static GWR network with the same number of neurons~\cite{Parisi2018c}.
While the mechanisms of structural plasticity in the Gamma-GWR do not resemble biologically plausible mechanisms (e.g.,~\cite{Ming2011}), the GWR learning algorithm represents an efficient computational model that incrementally adapts to non-stationary input.
Crucially, the GWR model creates new neurons whenever they are required and only after the training of existing ones.
The neural update rate decreases as the neurons become more habituated, which has the effect of preventing that noisy input interferes with consolidated neural representations.
Similar GWR-based approaches have been proposed for the incremental learning of body motion patterns~\cite{Mici2017, Elfaramawy2017, Parisi2016} and human-object interaction~\cite{Mici2018}.

To achieve invariance to scale and translation, we implemented MAX-pooling layers after each Gamma-GWR network with the receptive field of neurons increasing along the hierarchy.
This approach has shown competitive results with batch learning methods on the Weizmann~\cite{Gorelick2005} and the KTH~\cite{Schuldt2004} action benchmark datasets.
On the KTH dataset, this method outperforms the overall accuracy reported by \cite{jung2015} with three different deep learning models: convolutional neural network (CNN, 92.9\%), multiple spatiotemporal scales neural network (MSTNN, 95.3\%), and 3D-CNN (96.2\%).
A direct comparison with these methods is however hindered by the fact that they differ in the type of input and number of frames per sequence used during the training and the test phase.
On the Weizmann dataset, the Gamma-GWR model outperforms other hierarchical models that do not rely on handcrafted features, such as 3D CNN ($90.2\%$, \cite{Ji2013}) and 3D CNN in combination with long short-term memory ($94.39\%$, \cite{Baccouche2011}).

\begin{figure*}[t]
\centering
\includegraphics[width=0.4\textwidth]{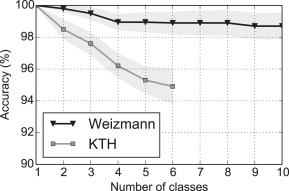}
\caption{Incremental learning: classification accuracy averaged across all the combinations for a given number of action classes~\cite{Parisi17}.}
\label{fig:ggwrr}
\end{figure*}

Overall, the Gamma-GWR model shows competitive results with batch learning methods on action benchmark datasets.
However, the neural growth and update are driven by the minimization of the bottom-up reconstruction error and, thus, without taking into account top-down, task-relevant signals that can regulate the plasticity-stability balance.
The mechanism to prevent the acquisition of new information from forgetting existing knowledge is embedded in the dynamics of the self-organizing learning algorithm that allocates new neurons or updates existing ones based on the discrepancy between the input distribution and the prototype neural weights.
To support this claim, we conducted an additional experiment in which we explore how our model accounts for avoiding catastrophic interference when learning new action classes.

For both datasets, we first trained the model with a single action class and then scaled up progressively to all the others in order to observe how the performance of the model changes for an increasing number of action classes. The results are shown in Fig.~\ref{fig:ggwrr}, where the classification accuracy was averaged across all the combinations for a given number of action classes.
Although the performance decreases as the number of action classes is increased, this decline is not catastrophic.
However, it is complex to establish whether this accuracy decrease is caused by catastrophic interference or the labeling strategy.
We observed that the overall quantization error of the networks tends to decrease over the training epochs, suggesting that the prototype neurons are effectively allocated and fine-tuned to better represent the input distribution.

\subsection{Replay via Neural Re-Activations}

This vanilla Gamma-GWR model has a set of limitations in OCL scenarios.
In particular, it is non-trivial to empirically find the adequate hyper-parameters to achieve an optimal plasticity-stability balance.
To completely alleviate catastrophic forgetting, the model may grow until it becomes computationally expensive to process data in real time, reducing its ability to work with large-scale datasets.
Instead, if the model excessively limits its growth, its overall performance over time may decrease due to progressive forgetting as it can be seen in Fig.~\ref{fig:ggwrr}.
A replay mechanism could prevent the model from progressive forgetting.
Additionally, although the networks would keep learning in an unsupervised fashion, a teaching signal can be used to modulate their growth during the learning phase.

\begin{figure*}[t]
\centering
\includegraphics[width=0.79\textwidth]{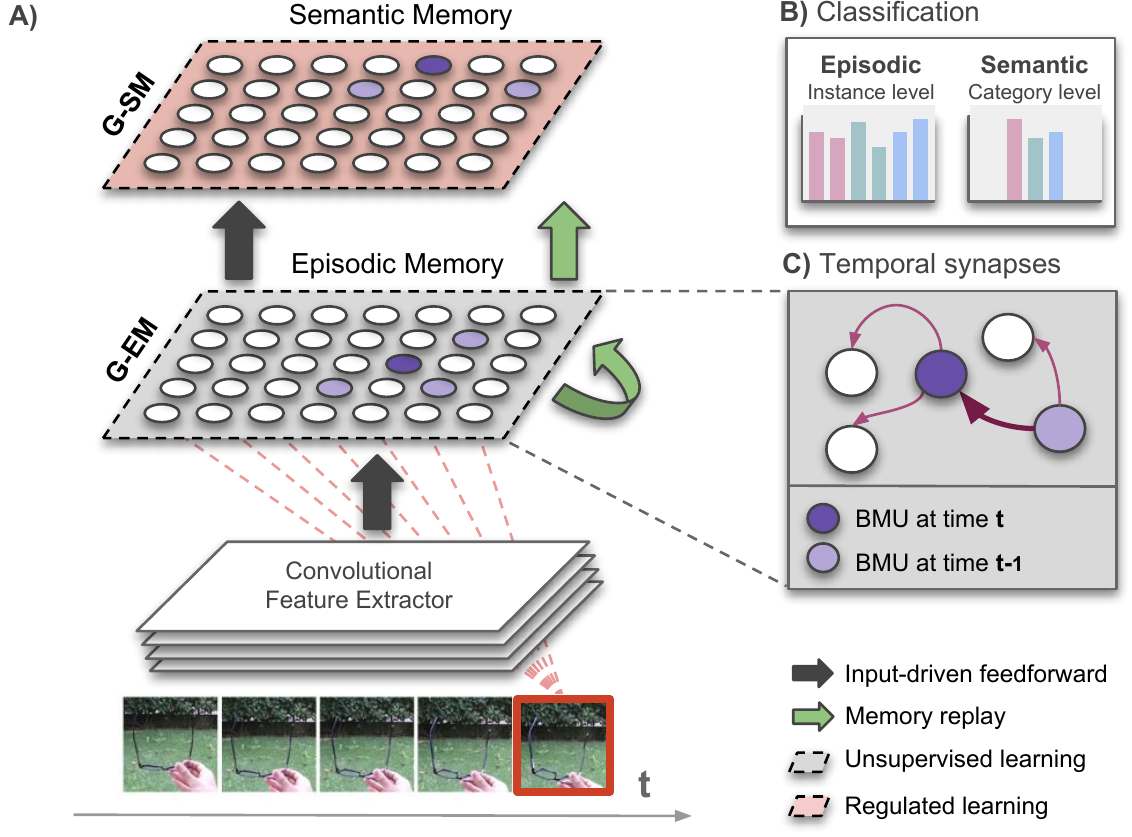}
\caption{(A) Illustration of the GDM architecture. Extracted features from image sequences are fed into a growing episodic memory (G-EM). Neural activation trajectories from G-EM are fed to the growing semantic memory (G-SM). While the learning process of G-EM remains unsupervised, G-SM uses class labels as task-relevant signals to regulate levels of plasticity. After each learning episode, neural re-activation trajectories are replayed to both memories (green arrows); (B) The architecture classifies image sequences at instance level
(episodic experience) and at category level (semantic knowledge). For classification, neurons in G-EM and G-SM associatively learn histograms of class labels from the input (red dashed lines); (C) To enable memory replay in the absence of sensory input, G-EM is equipped with temporal synapses that are strengthened between consecutively activated neurons~\cite{Parisi18}.}
\label{fig:architectureg}
\end{figure*}

In~\cite{Parisi18}, we proposed growing dual-memory learning (GDM) comprising a deep convolutional feature extractor and two hierarchically arranged recurrent self-organizing networks (Fig.~\ref{fig:architectureg}).
The GDM model comprises an episodic memory and a semantic memory, both implemented as extended Gamma-GWR networks~\cite{Parisi17}.
The growing episodic memory (G-EM) learns from sensory experience in an unsupervised fashion, i.e., levels of structural plasticity are regulated by the ability of the network to predict the spatiotemporal patterns given as input. Instead, the growing semantic memory (G-SM) receives neural activation trajectories from G-EM and uses task-relevant signals (input annotations) to modulate levels of neurogenesis and neural update, thereby developing more compact representations of statistical regularities.

G-EM and G-SM mitigate catastrophic forgetting through self-organizing learning
dynamics with structural plasticity, increasing information storage capacity in response to novel input.
For the consolidation of knowledge over time, internally generated neural activity patterns in G-EM are periodically replayed to both memories, thereby mitigating catastrophic forgetting during incremental learning tasks.
To yield periodic experience replay, G-EM is equipped with synapses that learn statistically significant neural activity in the temporal domain.
This results in the generation of sequence-selective neural activation trajectories that can be replayed to both networks after each learning episode without requiring additional memory resources to store the input.
To preserve the temporal structure of the input during experience replay, the model generates temporally-ordered trajectories of neural activity.
The trajectories are created by using the asymmetric temporal links of G-EM to recursively reactivate sequence-selective neural activity trajectories (RNATs) embedded in the network.
RNATs can be computed for each neuron in G-EM for a given temporal window and replayed to G-EM and G-SM after each learning episode triggered by external input stimulation.
For each neuron $j$ in G-EM, a RNAT $S_j$ of length $\lambda=K^{\text{EM}}+K^{\text{SM}}+1$ can be generated:
\begin{equation}\label{eq:RNATs}
S_j=\langle \textbf{w}^{\text{EM}}_{\text{s}(0)},\textbf{w}^{\text{EM}}_{\text{s}(1)},...,\textbf{w}^{\text{EM}}_{\text{s}(\lambda)} \rangle,
\end{equation}
\begin{equation}\label{eq:RNATs1}
\text{s}(i) = \arg\max_{n \in A \setminus {j}} P_{(n,\text{s}(i-1))}, i \in [1,\lambda],
\end{equation}
where $P_{(i,j)}$ is the matrix of temporal synapses and $\text{s}(0)=j$.

The set of generated RNATs from all G-EM neurons is replayed to G-EM and G-SM after each learning episode.
Sequence-selective prototype sequences can be generated and periodically replayed without the need to explicitly store the temporal relations and labels of previously seen training samples.
This is in agreement with neurophysiological studies evidencing that hippocampal replay consists of the reactivation of previously stored patterns of neural activity occurring predominantly after an experience~\cite{Kudrimoti4090,Karlsson2009}.

\begin{figure*}[t]
\centering
\includegraphics[width=0.8\textwidth]{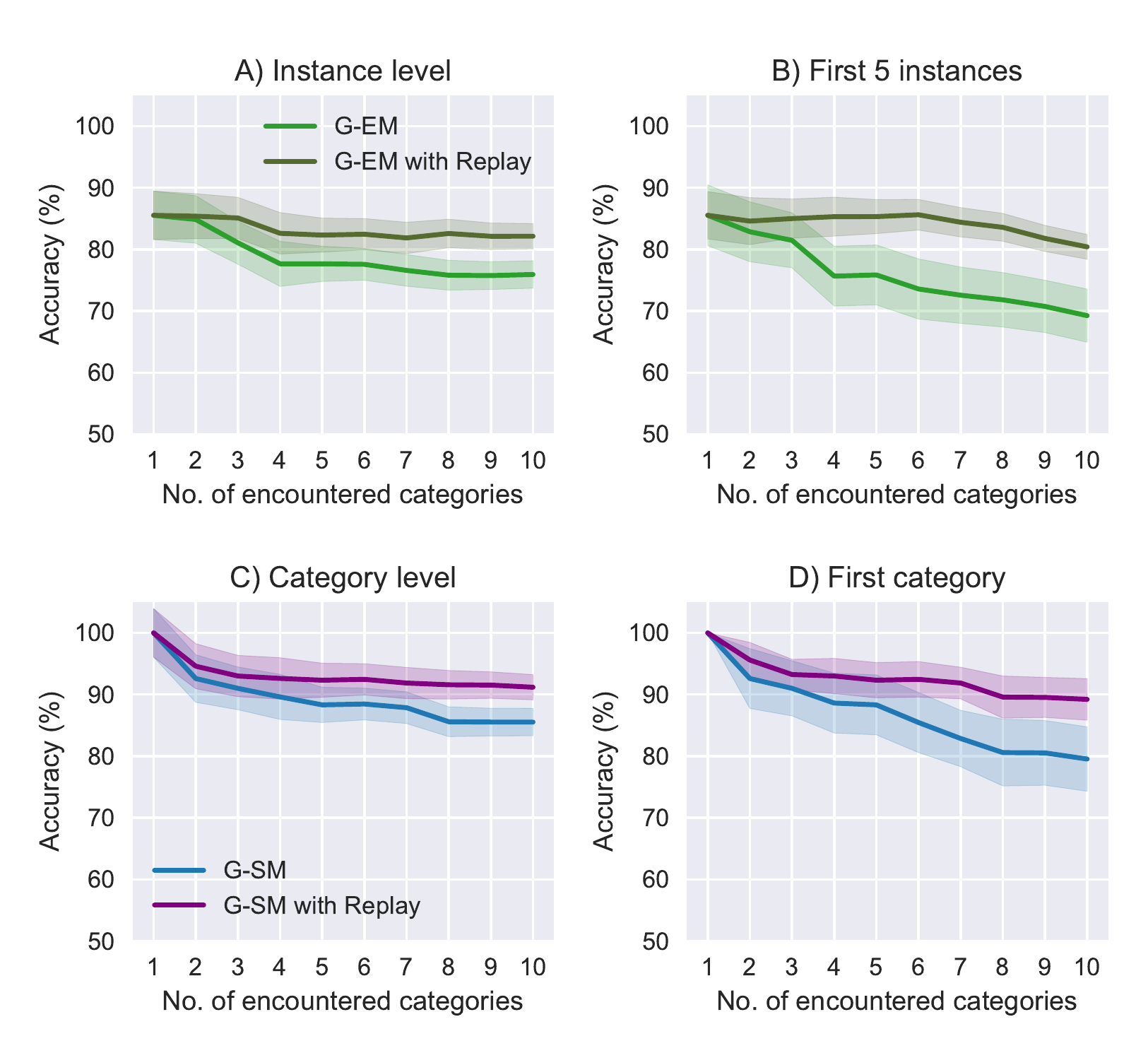}
\caption{Comparison of the effects of forgetting during incremental learning with and without memory replay at an instance level (A,B) and category level (C,D).
Each category contains 5 instances. The plots show the average accuracies on the categories encountered so far (A,C) and the accuracies on the first encountered
category (B,D) as further new categories are learned. The shaded areas show the standard deviation~\cite{Parisi18}.}
\label{fig:core50replay}
\end{figure*}

A series of experiments evaluating the performance of the GDM model on the CORe50 dataset show the benefit of using memory replay.
In the training strategy with replay, after each learning episode (i.e., a training epoch over the mini-batch), the model generates a set of RNATs $S_j$ from the G-EM neurons.
Thus, the number of RNATs of length $\lambda = 5$ is equal to the number of neurons created by G-EM.
RNATs are replayed to G-EM and G-SM in correspondence of novel sensory experience to reinforce previously encountered categories. 
Fig.~\ref{fig:core50replay} shows a comparison of the overall accuracy on all the object categories encountered so far to the accuracy on the first encountered category over the number of encountered categories.
At an instance level~(Fig.~\ref{fig:core50replay}.A-B), incremental learning with memory replay improves the overall accuracy to $82.14\%$ (from $75.93\%$) and accuracy on the first 5 instances to $80.41\%$ (from $69.25\%$).
At a category level~(Fig.~\ref{fig:core50replay}.C-D), the overall accuracy increases to $91.18\%$ (from $85.53\%$) and the accuracy on the first encountered category to $89.21\%$ (from $79.53\%$).
Overall, results show that replaying RNATs generated from G-EM significantly mitigates the effects of catastrophic forgetting over time.

\section{Conclusions}
\label{sec:conclusions}

OCL is a crucial but challenging aspect of learning systems.
Empirical evidence shows that the use of episodic memory can significantly alleviate catastrophic forgetting and that, when this memory replays latent input representations rather than explicitly stored training samples encountered during the learning phase, training complexity can be reduced.
Overall, generative memory replay strategies may not only empirically useful but also fundamental, especially when processing non-i.i.d. sequential data streams.

Future research should aim at modeling and integrating additional aspects of learning found in biological systems which would allow artificial agents to learn efficiently in more complex environments.
Examples of these aspects include \textit{curriculum} and \textit{transfer} learning.
Humans and animals exhibit better learning performance when training examples are organized in a meaningful way, e.g., by making the learning tasks gradually more difficult~\cite{Krueger2009}.
Following this observation, it was shown in~\cite{Elman1993} that having a curriculum of progressively harder tasks leads to faster training performance in neural
networks.
While a number of methods have been proposed that explore the use of training curricula~(e.g.~\cite{Graves2016, Reed2015}), these methods have not been investigated in combination with continual learning.
In transfer learning, previously acquired knowledge in one domain is applied to solve a problem in a novel domain~\cite{Holyoak1997}.
For this reason, transfer learning represents a significantly valuable feature for inferring general laws from (a limited amount of) samples.
Recent methods proposed, for instance, the use of growing neural networks to transfer learned low-level features and high-level policies from a simulated to a real environment~(e.g.~\cite{Rusu2017}).
However, scalable OCL systems with transferable knowledge remain an open and exciting challenge.

\subsection*{Acknowledgements}
The authors would like to thank the \emph{ContinualAI} organization and the other \emph{ContinualAI Research} fellows for their support.

\nocite{}
\bibliographystyle{spmpsci}
\bibliography{biblio.bib}

\begin{thebibliography}{10}
\providecommand{\url}[1]{{#1}}
\providecommand{\urlprefix}{URL }
\expandafter\ifx\csname urlstyle\endcsname\relax
  \providecommand{\doi}[1]{DOI~\discretionary{}{}{}#1}\else
  \providecommand{\doi}{DOI~\discretionary{}{}{}\begingroup
  \urlstyle{rm}\Url}\fi

\bibitem{Aimone2009}
Aimone, J.B., Wiles, J., Gage, F.H.: Computational influence of adult
  neurogenesis on memory encoding.
\newblock Neuron \textbf{61}, 187--202 (2009)

\bibitem{Baccouche2011}
Baccouche, M., Mamalet, F., Wolf, C., Garcia, C., Baskurt, A.: Sequential deep
  learning for human action recognition.
\newblock In: A.A. Salah, B.~Lepri (eds.) Human Behavior Understanding, pp.
  29--39. Springer Berlin Heidelberg, Berlin, Heidelberg (2011)

\bibitem{Bellemare13}
Bellemare, M.G., Naddaf, Y., Veness, J., Bowling, M.: The arcade learning
  environment: An evaluation platform for general agents.
\newblock Journal of Artificial Intelligence Research \textbf{47}, 253--279
  (2013)

\bibitem{Borji2016}
Borji, A., Izadi, S., Itti, L.: {iLab-20M: A Large-Scale Controlled Object
  Dataset to Investigate Deep Learning}.
\newblock In: International Conference of Computer Vision and Pattern
  Recogniton (CVPR), pp. 2221--2230 (2016)

\bibitem{Chen18}
Chen, Z., Liu, B.: Lifelong machine learning.
\newblock Synthesis Lectures on Artificial Intelligence and Machine Learning
  \textbf{12}(3), 1--207 (2018)

\bibitem{Deng2010}
Deng, W., Aimone, J.B., Gage, F.H.: New neurons and new memories: how does
  adult hippocampal neurogenesis affect learning and memory?
\newblock Nature Reviews Neuroscience \textbf{11}(5), 339--350 (2010)

\bibitem{Diaz18}
D{\'i}az-Rodr{\'i}guez, N., Lomonaco, V., Filliat, D., Maltoni, D.: {Don't
  forget, there is more than forgetting: new metrics for Continual Learning}.
\newblock In: {Workshop on Continual Learning, NeurIPS 2018 (Neural Information
  Processing Systems}. Montreal, Canada (2018)

\bibitem{Elfaramawy2017}
Elfaramawy, N., Barros, P., Parisi, G.I., Wermter, S.: Emotion recognition from
  body expressions with a neural network architecture.
\newblock pp. 143--149. Proceedings of the International Conference on Human
  Agent Interaction (HAI'17), Bielefeld, Germany (2017)

\bibitem{Elman1993}
Elman, J.L.: Learning and development in neural networks: The importance of
  starting small.
\newblock Cognition \textbf{48}(1), 71--99 (1993)

\bibitem{Everingham15}
Everingham, M., Eslami, S.M.A., Van~Gool, L., Williams, C.K.I., Winn, J.,
  Zisserman, A.: The pascal visual object classes challenge: A retrospective.
\newblock International Journal of Computer Vision \textbf{111}(1), 98--136
  (2015)

\bibitem{Fan18}
Fan, L., Zhu, Y., Zhu, J., Liu, Z., Zeng, O., Gupta, A., Creus-Costa, J.,
  Savarese, S., Fei-Fei, L.: Surreal: Open-source reinforcement learning
  framework and robot manipulation benchmark.
\newblock In: Conference on Robot Learning (2018)

\bibitem{Franco2009}
Franco, A., Maio, D., Maltoni, D.: {The Big Brother Database : Evaluating Face
  Recognition in Smart Home Environments}.
\newblock In: Advances in Biometrics: Third International Conference (ICB), pp.
  142--150 (2009)

\bibitem{French1999}
French, R.M.: Catastrophic forgetting in connectionist networks.
\newblock Trends in Cognitive Sciences \textbf{3}(4), 128--135 (1999)

\bibitem{Fusi2005}
Fusi, S., Drew, P.J., Abbott, L.F.: Cascade models of synaptically stored
  memories.
\newblock Neuron \textbf{45}(4), 599--611 (2005)

\bibitem{Gepperth16}
Gepperth, A., Hammer, B.: {Incremental learning algorithms and applications}.
\newblock In: {European Symposium on Artificial Neural Networks (ESANN)}.
  Bruges, Belgium (2016)

\bibitem{aloi}
Geusebroek, J.M., Burghouts, G.J., Smeulders, A.W.: {The Amsterdam Library of
  Object Images}.
\newblock International Journal of Computer Vision \textbf{61}(1), 103--112
  (2005)

\bibitem{Gorelick2005}
Gorelick, L., Blank, M., Shechtman, E., Irani, M., Basri, R.: Actions as
  space-time shapes.
\newblock pp. 1395--1402. ICCV'05, Beijing, China (2005)

\bibitem{Graves2016}
Graves, A., Wayne, G., Reynolds, M., Harley, T., Danihelka, I.,
  Grabska-Barwinska, A., Colmenarejo, S.G., Grefenstette, E., Ramalho, T.,
  Agapiou, J.e.a.: Hybrid computing using a neural network with dynamic
  external memory.
\newblock Nature \textbf{538}, 471--476 (2016)

\bibitem{Grossberg1980}
Grossberg, S.: How does a brain build a cognitive code?
\newblock Psychol. Rev. \textbf{87}, 1--51 (1980)

\bibitem{Hayes18MemoryEfficient}
Hayes, T.L., Cahill, N.D., Kanan, C.: Memory efficient experience replay for
  streaming learning.
\newblock 2019 International Conference on Robotics and Automation (ICRA) pp.
  9769--9776 (2018)

\bibitem{Hayes18NewMetrics}
Hayes, T.L., Kemker, R., Cahill, N.D., Kanan, C.: New metrics and experimental
  paradigms for continual learning.
\newblock In: 2018 IEEE/CVF Conference on Computer Vision and Pattern
  Recognition Workshops (CVPRW), pp. 2112--21123 (2018)

\bibitem{Holyoak1997}
Holyoak, K., Thagard, P.: The analogical mind.
\newblock American Psychologist \textbf{52}, 35--44 (1997)

\bibitem{loffe2017}
Ioffe, S.: {Batch Renormalization: Towards Reducing Minibatch Dependence in
  Batch-Normalized Models}.
\newblock In: Advances in neural information processing systems (NIPS), pp.
  1945----1953 (2017)

\bibitem{Ji2013}
{Ji}, S., {Xu}, W., {Yang}, M., {Yu}, K.: 3d convolutional neural networks for
  human action recognition.
\newblock IEEE Transactions on Pattern Analysis and Machine Intelligence
  \textbf{35}(1), 221--231 (2013)

\bibitem{jung2015}
Jung, M., Hwang, J., Tani, J.: Self-organization of spatio-temporal hierarchy
  via learning of dynamic visual image patterns on action sequences.
\newblock PLOS ONE \textbf{10}(7), 1--16 (2015)

\bibitem{Karlsson2009}
Karlsson, M., Frank, L.: Awake replay of remote experiences in the hippocampus.
\newblock Nature Neuroscience \textbf{19}(10), 913--918 (2009)

\bibitem{kemker18fearnet}
Kemker, R., Kanan, C.: Fearnet: Brain-inspired model for incremental learning.
\newblock In: International Conference on Learning Representations (2018)

\bibitem{Kemker17}
Kemker, R., McClure, M., Abitino, A., Hayes, T.L., Kanan, C.: Measuring
  catastrophic forgetting in neural networks.
\newblock In: AAAI (2017)

\bibitem{Kirkpatrick17}
Kirkpatrick, J., Pascanu, R., Rabinowitz, N., Veness, J., Desjardins, G., Rusu,
  A.A., Milan, K., Quan, J., Ramalho, T., Grabska-Barwinska, A., et~al.:
  Overcoming catastrophic forgetting in neural networks.
\newblock Proc. of the national academy of sciences  (2017)

\bibitem{Krizhevsky09}
Krizhevsky, A., Hinton, G., et~al.: Learning multiple layers of features from
  tiny images.
\newblock Tech. rep., Citeseer (2009)

\bibitem{krizhevsky12}
Krizhevsky, A., Sutskever, I., Hinton, G.E.: Imagenet classification with deep
  convolutional neural networks.
\newblock In: Advances in neural information processing systems, pp. 1097--1105
  (2012)

\bibitem{Krueger2009}
Krueger, K.A., Dayan, P.: Flexible shaping: how learning in small steps helps.
\newblock Cognition \textbf{110}, 380--394 (2009)

\bibitem{Kudrimoti4090}
Kudrimoti, H.S., Barnes, C.A., McNaughton, B.L.: Reactivation of hippocampal
  cell assemblies: Effects of behavioral state, experience, and eeg dynamics.
\newblock Journal of Neuroscience \textbf{19}(10), 4090--4101 (1999)

\bibitem{lake2015human}
Lake, B.M., Salakhutdinov, R., Tenenbaum, J.B.: Human-level concept learning
  through probabilistic program induction.
\newblock Science \textbf{350}(6266), 1332--1338 (2015)

\bibitem{LeCun10}
LeCun, Y., Cortes, C.: {MNIST} handwritten digit database.
\newblock public  (2010)

\bibitem{LeCun2004}
LeCun, Y., Huang, F.J., Bottou, L.: {Learning methods for generic object
  recognition with invariance to pose and lighting}.
\newblock In: Proceedings of the 2004 IEEE Computer Society Conference on
  Computer Vision and Pattern Recognition (CVPR), vol.~2, pp. 97--104 (2004)

\bibitem{lesort:hal-02381343}
Lesort, T., Lomonaco, V., Stoian, A., Maltoni, D., Filliat, D.,
  D{\'i}az-Rodr{\'i}guez, N.: {Continual Learning for Robotics: Definition,
  Framework, Learning Strategies, Opportunities and Challenges}.
\newblock {Information Fusion}  (2019)

\bibitem{Li17}
Li, Z., Hoiem, D.: Learning without forgetting.
\newblock IEEE Transactions on Pattern Analysis and Machine Intelligence
  (2017)

\bibitem{lomonaco2019continual}
Lomonaco, V., Desai, K., Culurciello, E., Maltoni, D.: Continual reinforcement
  learning in 3d non-stationary environments.
\newblock arXiv preprint arXiv:1905.10112  (2019)

\bibitem{Lomonaco2016}
Lomonaco, V., Maltoni, D.: {Comparing Incremental Learning Strategies for
  Convolutional Neural Networks}.
\newblock In: Artificial Neural Networks in Pattern Recognition: 7th IAPR TC3
  Workshop (ANNPR 2016), pp. 175--184 (2016)

\bibitem{Lomonaco17}
Lomonaco, V., Maltoni, D.: {CORe50: a New Dataset and Benchmark for Continuous
  Object Recognition}.
\newblock In: S.~Levine, V.~Vanhoucke, K.~Goldberg (eds.) Proceedings of the
  1st Annual Conference on Robot Learning, \emph{Proceedings of Machine
  Learning Research}, vol.~78, pp. 17--26. PMLR (2017)

\bibitem{pmlr-v78-lomonaco17a}
Lomonaco, V., Maltoni, D.: {CORe50: a New Dataset and Benchmark for Continuous
  Object Recognition}.
\newblock In: S.~Levine, V.~Vanhoucke, K.~Goldberg (eds.) Proceedings of the
  1st Annual Conference on Robot Learning, \emph{Proceedings of Machine
  Learning Research}, vol.~78, pp. 17--26. PMLR (2017)

\bibitem{Lomonaco2017-arxiv}
Lomonaco, V., Maltoni, D.: {CORe50: a New Dataset and Benchmark for Continuous
  Object Recognition}.
\newblock arXiv preprint arXiv:1705.03550  (2017)

\bibitem{Lomonaco2019}
Lomonaco, V., Maltoni, D., Pellegrini, L.: {Fine-Grained Continual Learning}.
\newblock arXiv preprint arXiv: 1907.03799 pp. 1--14 (2019)

\bibitem{Lopez-Paz17}
Lopez-Paz, D., Ranzato, M.A.: Gradient episodic memory for continual learning.
\newblock In: I.~Guyon, U.V. Luxburg, S.~Bengio, H.~Wallach, R.~Fergus,
  S.~Vishwanathan, R.~Garnett (eds.) Advances in Neural Information Processing
  Systems 30, pp. 6467--6476. Curran Associates, Inc. (2017)

\bibitem{Maltoni2016-icpr}
Maltoni, D., Lomonaco, V.: {Semi-supervised Tuning from Temporal Coherence}.
\newblock pp. 2509--2514 (2016)

\bibitem{Maltoni18}
Maltoni, D., Lomonaco, V.: Continuous learning in single-incremental-task
  scenarios.
\newblock Neural Networks \textbf{116}, 56 -- 73 (2019)

\bibitem{maltoni2019}
Maltoni, D., Lomonaco, V.: {Continuous learning in single-incremental-task
  scenarios}.
\newblock Neural Networks \textbf{116}, 56--73 (2019)

\bibitem{Mandlekar18}
Mandlekar, A., Zhu, Y., Garg, A., Booher, J., Spero, M., Tung, A., Gao, J.,
  Emmons, J., Gupta, A., Orbay, E., Savarese, S., Fei-Fei, L.: Roboturk: A
  crowdsourcing platform for robotic skill learning through imitation.
\newblock In: Conference on Robot Learning (2018)

\bibitem{Mankowitz18}
Mankowitz, D.J., {\v{Z}}{\'\i}dek, A., Barreto, A., Horgan, D., Hessel, M.,
  Quan, J., Oh, J., van Hasselt, H., Silver, D., Schaul, T.: Unicorn: Continual
  learning with a universal, off-policy agent.
\newblock arXiv preprint arXiv:1802.08294  (2018)

\bibitem{Marsland2002}
Marsland, S., Shapiro, J., Nehmzow, U.: A self-organising network that grows
  when required.
\newblock Neural Networks \textbf{15}(8--9), 1041--1058 (2002)

\bibitem{Mcclelland95}
McClelland, J.L., McNaughton, B.L., O'reilly, R.C.: Why there are complementary
  learning systems in the hippocampus and neocortex: insights from the
  successes and failures of connectionist models of learning and memory.
\newblock Psychological review \textbf{102}(3), 419 (1995)

\bibitem{Mermillod13}
Mermillod, M., Bugaiska, A., Bonin, P.: {The stability-plasticity dilemma:
  investigating the continuum from catastrophic forgetting to age-limited
  learning effects.}
\newblock Frontiers in psychology \textbf{4}(August), 504 (2013)

\bibitem{Mici2017}
Mici, L., Parisi, G.I., Wermter, S.: An incremental self-organizing
  architecture for sensorimotor learning and prediction.
\newblock arXiv:1712.08521 (2017)

\bibitem{Mici2018}
Mici, L., Parisi, G.I., Wermter, S.: A self-organizing neural network
  architecture for learning human-object interactions.
\newblock Neurocomputing \textbf{307}, 14--24 (2018)

\bibitem{Ming2011}
Ming, G.L., Song, H.: Adult neurogenesis in the mammalian brain: Significant
  answers and significant questions.
\newblock Neuron \textbf{70}, 687--702 (2011)

\bibitem{Nene1996}
Nene, S.A., Nayar, S.K., Murase, H.: {Columbia Object Image Library
  (COIL-100)}.
\newblock Technical Report  (1996)

\bibitem{Netzer11}
Netzer, Y., Wang, T., Coates, A., Bissacco, A., Wu, B., Ng, A.Y.: Reading
  digits in natural images with unsupervised feature learning.
\newblock In: NIPS Workshop on Deep Learning and Unsupervised Feature Learning
  2011 (2011)

\bibitem{Parisi2018c}
Parisi, G., Ji, X., Wermter, S.: On the role of neurogenesis in overcoming
  catastrophic forgetting.
\newblock NIPS'18, Workshop on Continual Learning, Montreal, Canada (2018)

\bibitem{Parisi18review}
Parisi, G.I., Kemker, R., Part, J.L., Kanan, C., Wermter, S.: Continual
  lifelong learning with neural networks: A review.
\newblock Neural Networks \textbf{113}, 54 -- 71 (2019)

\bibitem{Parisi2016}
Parisi, G.I., Magg, S., Wermter, S.: Human motion assessment in real time using
  recurrent self-organization.
\newblock pp. 71--79. Proceedings of the IEEE International Symposium on Robot
  and Human Interactive Communication, New York, NY (2016)

\bibitem{Parisi2017a}
Parisi, G.I., Tani, J., Weber, C., Wermter, S.: Lifelong learning of humans
  actions with deep neural network self-organization.
\newblock Neural Networks \textbf{96}, 137--149 (2017)

\bibitem{Parisi18}
Parisi, G.I., Tani, J., Weber, C., Wermter, S.: Lifelong learning of
  spatiotemporal representations with dual-memory recurrent self-organization.
\newblock Frontiers in Neurorobotics \textbf{12}, 78 (2018)

\bibitem{Parisi17}
Parisi, S., Ramstedt, S., J., P.: Goal-driven dimensionality reduction for
  reinforcement learning.
\newblock In: Proceedings of the IEEE/RSJ Conference on Intelligent Robots and
  Systems (IROS) (2017)

\bibitem{Pasquale2015a}
Pasquale, G., Ciliberto, C., Odone, F., Rosasco, L., Natale, L.: {Teaching iCub
  to recognize objects using deep Convolutional Neural Networks}.
\newblock In: Proceedings of Workshop on Machine Learning for Interactive
  Systems, pp. 21--25 (2015)

\bibitem{icub-transf}
Pasquale, G., Ciliberto, C., Rosasco, L., Natale, L.: {Object Identification
  from Few Examples by Improving the Invariance of a Deep Convolutional Neural
  Network}.
\newblock In: 2016 IEEE/RSJ International Conference on Intelligent Robots and
  Systems (IROS), pp. 4904--4911 (2016)

\bibitem{pellegrini2019latent}
Pellegrini, L., Graffieti, G., Lomonaco, V., Maltoni, D.: Latent replay for
  real-time continual learning.
\newblock arXiv preprint arXiv:1912.01100  (2019)

\bibitem{Rebuffi16}
{Rebuffi}, S., {Kolesnikov}, A., {Sperl}, G., {Lampert}, C.H.: icarl:
  Incremental classifier and representation learning.
\newblock In: 2017 IEEE Conference on Computer Vision and Pattern Recognition
  (CVPR), pp. 5533--5542 (2017)

\bibitem{Reed2015}
Reed, S., de~Freitas, N.: Neural programmer interpreters.
\newblock arXiv:1511.06279 (2015)

\bibitem{Richardson08}
Richardson, F.M., Thomas, M.S.: Critical periods and catastrophic interference
  effects in the development of self-organizing feature maps.
\newblock Developmental science \textbf{11}(3), 371--389 (2008)

\bibitem{Robins95}
Robins, A.: Catastrophic forgetting, rehearsal and pseudorehearsal.
\newblock Connection Science \textbf{7}(2), 123--146 (1995)

\bibitem{Rusu16progressive}
{Rusu}, A.A., {Rabinowitz}, N.C., {Desjardins}, G., {Soyer}, H., {Kirkpatrick},
  J., {Kavukcuoglu}, K., {Pascanu}, R., {Hadsell}, R.: {Progressive Neural
  Networks}.
\newblock ArXiv e-prints  (2016)

\bibitem{Rusu2017}
Rusu, A.A., Vecerik, M., Roth{\"o}rl, T., Heess, N., Pascanu, R., Hadsell, R.:
  Sim-to-real robot learning from pixels with progressive nets.
\newblock CoRL'17, Mountain View, CA (2017)

\bibitem{Schuldt2004}
Schuldt, C., Laptev, I., Caputo, B.: Recognizing human actions: A local {SVM}
  approach.
\newblock pp. 32--36. ICPR'04, Cambridge, UK (2004)

\bibitem{Schwarz2015}
Schwarz, M., Schulz, H., Behnke, S.: {RGB-D Object Recognition and Pose
  Estimation based on Pre-trained Convolutional Neural Network Features}.
\newblock IEEE International Conference on Robotics and Automation (ICRA'15)
  (May), 1329--1335 (2015)

\bibitem{openloris}
She, Q., Feng, F., Hao, X., Yang, Q., Lan, C., Lomonaco, V., Shi, X., Wang, Z.,
  Guo, Y., Zhang, Y., Qiao, F., Chan, R.H.M.: Openloris-object: {A} dataset and
  benchmark towards lifelong object recognition.
\newblock CoRR \textbf{abs/1911.06487} (2019)

\bibitem{Shin17}
Shin, H., Lee, J.K., Kim, J., Kim, J.: Continual learning with deep generative
  replay.
\newblock In: Advances in Neural Information Processing Systems, pp. 2990--2999
  (2017)

\bibitem{bigbird}
Singh, A., Sha, J., Narayan, K.S., Achim, T., Abbeel, P.: {BigBIRD: A
  Large-Scale 3D Database of Object Instances}.
\newblock In: 2014 IEEE International Conference on Robotics and Automation
  (ICRA), pp. 509--516 (2014)

\bibitem{Welinder10}
Welinder, P., Branson, S., Mita, T., Wah, C., Schroff, F., Belongie, S.,
  Perona, P.: {Caltech-UCSD Birds 200}.
\newblock Tech. Rep. CNS-TR-2010-001, California Institute of Technology (2010)

\bibitem{wu2018memory}
Wu, C., Herranz, L., Liu, X., wang, y., van~de Weijer, J., Raducanu, B.: Memory
  replay gans: Learning to generate new categories without forgetting.
\newblock In: S.~Bengio, H.~Wallach, H.~Larochelle, K.~Grauman,
  N.~Cesa-Bianchi, R.~Garnett (eds.) Advances in Neural Information Processing
  Systems 31, pp. 5962--5972. Curran Associates, Inc. (2018)

\bibitem{Xiao2017}
Xiao, H., Rasul, K., Vollgraf, R.: Fashion-mnist: a novel image dataset for
  benchmarking machine learning algorithms.
\newblock arXiv preprint arXiv:1708.07747  (2017)

\bibitem{Yu15}
Yu, F., Zhang, Y., Song, S., Seff, A., Xiao, J.: Lsun: Construction of a
  large-scale image dataset using deep learning with humans in the loop.
\newblock CoRR \textbf{abs/1506.03365} (2015)

\bibitem{Zenke17}
{Zenke}, F., {Poole}, B., {Ganguli}, S.: Continual learning through synaptic
  intelligence.
\newblock In: D.~Precup, Y.W. Teh (eds.) Proceedings of the 34th International
  Conference on Machine Learning, \emph{Proceedings of Machine Learning
  Research}, vol.~70, pp. 3987--3995. PMLR, International Convention Centre,
  Sydney, Australia (2017)

\end{thebibliography}
\end{document}